\pdfoutput=1
\documentclass[11pt]{article}
\usepackage{acl}
\usepackage{times}
\usepackage{latexsym}
\usepackage{booktabs} 
\usepackage[T1]{fontenc}
\usepackage[utf8]{inputenc}
\usepackage{microtype}
\usepackage{inconsolata}
\usepackage{graphicx}
\usepackage{tabularx}
\usepackage{algorithm}
\usepackage{algpseudocode}
\usepackage{multirow}
\usepackage{amsmath}
\title{Building a Stable Planner: An Extended Finite State Machine Based Planning Module for Mobile GUI Agent}

\author{Fanglin Mo\(^{1,2}\), Junzhe Chen\(^1\), Haoxuan Zhu\(^2\), Xuming Hu\(^1\)\thanks{Corresponding author}\\
        \(^1\)Hong Kong University of Science and Technology (Guangzhou) \\ 
        \(^2\)School of Computer Science \& Engineering, South China University of Technology}

\newcommand{\figref}[1]{Fig.~\ref{#1}}
\newcommand{\tabref}[1]{Tab.~\ref{#1}}

\newcommand{\eqnref}[1]{Eq.~\ref{#1}}
\newcommand{\alref}[1]{Algorithm~\ref{#1}}

\begin{document}
\maketitle
\begin{abstract}
Mobile GUI agents execute user commands by directly interacting with the graphical user interface (GUI) of mobile devices, demonstrating significant potential to enhance user convenience. However, these agents face considerable challenges in task planning, as they must continuously analyze the GUI and generate operation instructions step by step. This process often leads to difficulties in making accurate task plans, as GUI agents lack a deep understanding of how to effectively use the target applications, which can cause them to become "lost" during task execution. To address the task planning issue, we propose SPlanner, a plug-and-play planning module to generate execution plans that guide vision language model(VLMs) in executing tasks. The proposed planning module utilizes extended finite state machines (EFSMs) to model the control logits and configurations of mobile applications. It then decomposes a user instruction into a sequence of primary function modeled in EFSMs, and generate the execution path by traversing the EFSMs. We further refine the execution path into a natural language plan using an LLM. The final plan is concise and actionable, and effectively guides VLMs to generate interactive GUI actions to accomplish user tasks. SPlanner demonstrates strong performance on dynamic benchmarks reflecting real-world mobile usage. On the AndroidWorld benchmark, SPlanner achieves a 63.8\% task success rate when paired with Qwen2.5-VL-72B as the VLM executor, yielding a 28.8 percentage point improvement compared to using Qwen2.5-VL-72B without planning assistance.
\end{abstract}

\section{Introduction}
\par The LLM-brained mobile graphic user interface (GUI) agent is designed to help users control their mobile devices using natural language (hereafter referred to simply as GUI agent). It uses a large language model(LLM) to interpret user instructions and execute tasks step by step. At each step, the agent analyzes the current GUI state and generates an operation command that simulates human interaction. Unlike agents that rely on predefined scripts or APIs, GUI agent makes decisions based on its understanding of the GUI, enabling it to handle more diverse user commands and complex interface scenarios. Moreover, by generating operation commands that mimic human interactions, the GUI agent can bypass restrictions imposed by the software permissions of mobile devices. While such a vision of GUI agent promises to significantly enhance the convenience of mobile device usage, their performance still falls short of practical deployment.
\par One of the major challenges facing GUI agent is task planning, especially for high-level tasks and unseen interfaces. Planning involves anticipating outcomes and choosing optimal paths. However, most current LLMs lack robust reasoning mechanisms on multi-step tasks. Moreover, they have limited understanding of how mobile applications are used, and tend to focus on screen elements that are only literally related to the tasks. As a result, current LLM-brained GUI agents are prone to getting stuck in recurring errors and need to repeatedly plan during task execution. We argue that equipping GUI agents with knowlege about the usage of mobile applications, such as their operational logic and critical functionalities, can facilitate the effectiveness and robustness of planning. With this knowledge, GUI agents know how to navigate through various applications and to cope with complex user requests.
\par Existing GUI agents acquire application usage knowledge from operation examples collected during task execution. 
These examples are either used to adjust models using training approaches like fine-tuning, or are integrated into knowledge bases to continuously improve the performance of agents. While these methods have yield positive results, they struggle with the problem of generalization, with their performance deteriorating significantly on unseen mobile applications. Besides, modern applications are frequently updated, and previous knowledge may become invalid. To adapt to new application versions, both retraining and updating the knowledge bases are too cost-ineffective to be adopted.
\par To equip GUI agents with knowledge of application usage more effectively and flexibly, we propose to model the paths users take within an application using Extended Finite State Machines (EFSMs). The EFSMs for various applications are then utilized by a plug-and-play planning module to facilitate the planning performance of GUI agents. EFSM extends traditional finite state machines by incorporating not only the standard components—states, events, output actions and transitions—but also variables and guard conditions, and , which enable it to represent more complex and conditional interaction flows as well as internal status. In our modeling approach, each page of a mobile application is represented as a state, and the primary functions of the application are defined as output actions. We then construct the state transitions based on the application’s structural layout and operational logic. At runtime, given the target primary functions, the EFSM is traversed to identify a complete execution path from the initial state (page) to the output action(primary function). This execution path is highly interpretable and stable, and is further refined into a natural language plan to guide the agent in completing the task.

\par In this paper, we propose SPlanner, an EFSM-based planning module designed to generate execution plans that assist GUI agents in accomplishing user tasks. The proposed planning module operates in three steps. Upon receiving a user instruction, SPlanner first uses a LLM (e.g., Deepseek V3) to parse the instruction, identifying the target applications and critical functionalities related to the user's task. Then, the EFSMs corresponding to the applications is solved to generate a task execution path that satisfies the specified actions. Finally, this path is polished and refined based on the user’s intent, yielding a natural-language task execution plan. Once the plan is generated, the VLM incrementally produces interactive actions(e.g., click, swipe) by combining the plan with the visual understanding of the current screen. We evaluate our method on AndroidWorld, a dynamic benchmark that closely mirrors real-world mobile application scenarios. Using the off-the-shelf generalist VLM Qwen2.5-VL-72B as the executor, our approach achieves a task success rate of 63.8\%, representing a substantial improvement of 28.8 percentage points over the baseline performance of Qwen2.5-VL-72B without our planning module.
\par In summary, our main contributions are as follows:
\begin{itemize}
    \item We propose a novel approach to modeling mobile applications using EFSMs, enabling GUI agents to acquire application usage knowledge in a direct and interpretable manner.
    \item We introduce SPlanner, an EFSM-based planning module to stably generate reliable and structured task execution plans.
    \item We perform comprehensive evaluations of SPlanner on a dynamic and realistic benchmarks, AndroidWorld demonstrating its effectiveness in executing user instructions.
\end{itemize}

\section{Related work}

%***********************Section 2-1***********************

\subsection{LLM-brained GUI agent}
\par Recently, LLM-powered mobile GUI agents have garnered significant attention, leading to the emergence of various novel agents, including some cross-platform solutions that work across mobile devices, web browsers, and computers. In the context of mobile GUI agents, the contributions of these methods can be broadly categorized into two key areas: grounding and reasoning.
\par To improve the agent’s grounding ability in GUI environments, several approaches have been proposed. Methods such as UGround \cite{Uground}, UI-TARS \cite{UI-TARS}, and SeeClick \cite{seeclick} leverage efficient fine-tuning techniques combined with high-quality training data to enhance the model’s understanding of GUI elements. Other approaches simplify the grounding process by incorporating OCR or GUI XML files—for example, COCO-Agent \cite{coco-agent} uses OCR to generate bounding boxes around GUI components, allowing the agent to select them more effectively. Additionally, some methods apply prompt engineering to boost grounding performance; for instance, CoAT \cite{coat} adopts Chain-of-Thought (CoT) prompting, guiding the agent to first generate a textual description of the GUI, thereby encouraging deeper semantic understanding of interface elements.
\par In terms of reasoning, the core objective is to enable the agent to master how to operate mobile applications. Some approaches, such as SeeAct \cite{seeact}, leverage Self-Reflection techniques, allowing the agent to learn from its own mistakes and progressively improve its application usage capabilities. Others, like AutoDroid \cite{autodroid} and MobileGPT \cite{mobilegpt}, adopt Self-Evolution strategies, where a knowledge base is built from the agent’s interaction history to support decision-making and reasoning. Additionally, certain methods focus on constructing new training datasets to facilitate learning. For instance, CoAT \cite{coat} introduces a dataset that not only contains GUI screenshots and corresponding actions, but also includes detailed action analyses and their outcomes, providing richer context for the agent to learn from.
\par Although existing methods have made notable progress in addressing the reasoning challenge that this work focuses on, they often rely on extensive data collection or costly training processes. Moreover, these approaches still struggle to achieve sufficient stability and interpretability. This motivates our focus on leveraging symbolic systems as a more transparent and cost-effective solution to the reasoning problem.

%***********************Section 2-2***********************

\subsection{Symbolic Planner in LLM-brained agent}
\par To address the planning challenge of LLM-powered agents, one approach is to introduce an additional planning module responsible for defining the plan, and a common strategy is to use a Symbolic Planner, which relies on a well-established symbolic model to represent the problem and employs symbolic reasoning to determine the optimal path from the initial state to the target state. A representative example is LLM+P \cite{llm+p}, which utilizes a Symbolic Planner based on the Planning Domain Definition Language(PDDL) model. In this approach, the LLM parses the problem into the PDDL format, and a solver is then used to find the best path by solving the formalized problem. Another notable work, LLM+ASP \cite{llm+asp}, employs a Symbolic Planner based on Answer Set Programming (ASP), where the LLM converts the problem into an ASP-compatible format, and an ASP solver is used to determine the task path. These methods use symbolic system solvers to complete path reasoning, making the path solving process extremely stable and explainable.
\par Although the aforementioned Symbolic Planner avoids requiring the LLM to perform logical reasoning to generate a plan, it still necessitates that the LLM models the entire problem (or understands it as a scenario) within a symbolic system and describes the task in a formal language. However, real-world problems are often complex and dynamic, making it rare for LLMs to model the entire problem accurately, which significantly limits the applicability of these Symbolic Planners. One possible approach is to manually model the task-related problem into a symbolic system before execution, though this demands significant expertise and effort from human experts. Nevertheless, in the context of mobile application scenarios, the cost of manually modeling the application into a symbolic system becomes more feasible. Given this, we propose combining the Symbolic Planner with a mobile GUI agent to address these challenges.

%***********************Section 3-0***********************

\section{Method}
\par In this section, we detail the workflow of SPlanner. We first introduce EFSMs, which are used to model applications and construct a structured knowledge base. Based on this knowledge, we build a planning module and employ a VLM as the executor. During task execution, the planning module first generates a detailed task plan from the user's instructions, which the VLM then follows to complete the task step by step.

%***********************Figure: Big Graph***********************

\begin{figure*}[t]
  \centering
  \includegraphics[width=1.0\linewidth]{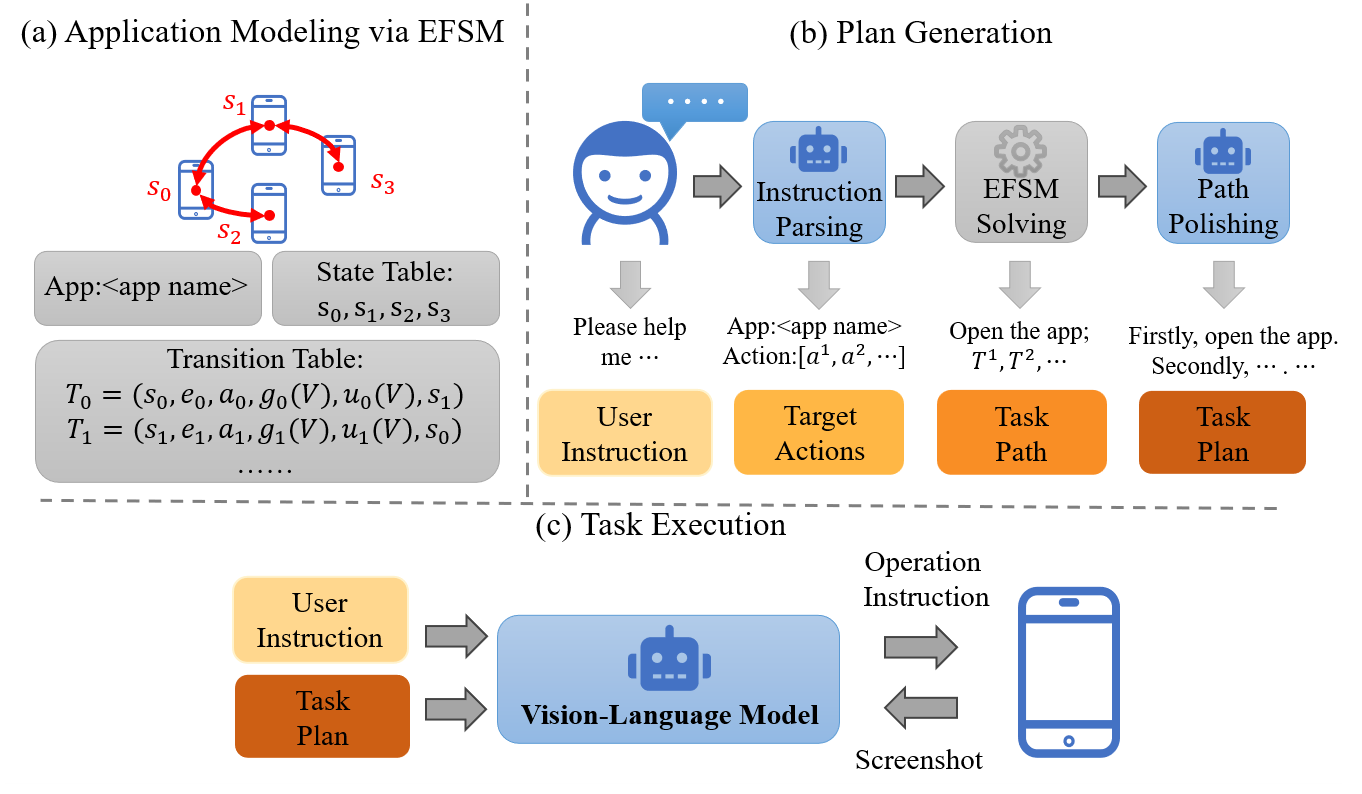}
  \caption{The SPlanner workflow consists of three main stages. First, Application Modeling via EFSM: (a) Prior to deployment, each target application is manually modeled into an EFSM, described using a set of state tables and state transition tables. Second, Plan Generation: (b) Upon receiving a user instruction, SPlanner processes it through three subprocedures — Instruction Parsing, EFSM Solving, and Path Polishing — to generate a detailed execution plan, with superscripts of \(a\) and \(T\) indicating their respective order of generation. Third, Task Execution with VLM: (c) We employs a VLM to execute the task by sequentially observing mobile device screenshots and following the generated plan, step by step, until the task is completed.}
  \label{fig:BigGraph}
\end{figure*}

%***********************Table: Symbol table***********************
\begin{table*}[t]

\renewcommand{\arraystretch}{1.3}
\begin{tabularx}{\textwidth}{cXX}
\toprule
\textbf{Symbol} & \textbf{Explanation} & \textbf{Examples} \\
\midrule
\(s\in S\) & A screen or page of the application & Camera home page; Camera settings page \\
\(e\in E\) & A sequence of user operations performed on the GUI & Click the button at the bottom of the screen, then click again after \{duration\} \\
\(a\in A\) & A primary function performed by the application & Take a photo; Record a video of \{duration\} \\
\(V\) & Internal variables describing the application’s configuration & Video mode = True; Front camera mode = False \\
\(g(V)\) & Guard conditions that must be satisfied for a transition to occur & if Video mode = True \\
\(u(V)\) & Update function applied to variables during a transition & Video mode \(\rightarrow\) False \\
\bottomrule
\end{tabularx}
\centering
\caption{Explanation of symbols in EFSM. The first column lists each transition component, while the second column explains its meaning in the context of mobile application modeling. Additionally, the third column provides representative examples for each component.}
\label{tab:symbol_explanation}
\end{table*}

%***********************Section 3-1***********************

\subsection{Mobile application modeling via EFSM}
\par To formally describe the behavior of a mobile application, we adopt the Extended Finite State Machine (EFSM) as our modeling framework. EFSM extend classical finite state machines by incorporating variables and guard conditions, thereby enabling the representation of both control logic and data-dependent behaviors within a unified formalism. Formally, in the scenario of application modeling, an EFSM is defined as a tuple:
\begin{equation}
    \varepsilon =(S,E,A,V,T,s_0).
\end{equation}
\par Here, \(S\) denotes the set of states, each representing a screen of the app, and \(s_0 \in S\) is the initial state, typically corresponding to the launch screen or entry point. \(A\)(referred to as the action set in the original EFSM) denotes the set of primary functions, which encapsulate the core functionalities or intended purposes of the application. \(V\) is the set of variables used to describe the app’s internal configuration. \(E\) denotes the set of events, each representing a sequence of operations performed on the graphical user interface (GUI) that trigger a transition in \(T\), and are commonly described in natural language. \(T\) is the set of transitions, each can be represented as \((s, e, a, g(V), u(V), s') \in T\), where \(s \in S\) and \(s' \in S\) are the source and target states, respectively; \(e \in E\) is the event; \(a \in A\) is the output primary function executed during the transition; \(u(V)\) denotes the update function that modifies the variables; and \(g(V)\) specifies the guard conditions that must be satisfied for the transition to occur. During a transition, the application may perform a primary function, update its internal variables, or navigate from one screen to another. For clarity and ease of understanding, we summarize the meaning of each EFSM component in \tabref{tab:symbol_explanation}
\par In SPlanner, we utilize EFSMs to model all the mobile applications involved, resulting in a structured knowledge base composed of multiple EFSM instances:

\begin{equation}
\label{eq:(2)}
\begin{split}
  &\mathcal F=\{\varepsilon _1,\varepsilon _2,\cdots,\varepsilon_n\}, \\
  &\varepsilon_j = (S^j,E^j,A^j,T^j,V^j,s_0^j), \\
  &j=1,2,\cdots,n.
\end{split}
\end{equation}
\par Each EFSM \(\varepsilon^j\) encapsulates the state-transition dynamics of a specific application. Given a sequence of target primary function that implements the user instruction,
\begin{equation}
A^T=(a_1,a_2,\cdots ,a_k) \subseteq A^j \label{eq:(3)}.
\end{equation}
a valid task execution path, starting from the initial state \(s_0^j \in S^j\) and invoking all the primary function in \(A^T\), can be derived by traversing the state machine \(\varepsilon^j\). We use search algorithms, such as Breadth-First Search (BFS), to compute a transition path:
\begin{equation}
\label{eq:(4)}
\begin{split}
&p=(t_1,t_2,\cdots,t_m), \\
&t _i=(s_{i-1},e_i,a_i,g_i(V),u_i(V),s_i) \in T^j .
\end{split}
\end{equation}
Here \(s_0\) represents the initial state \(s_0^j\), while \(s_i\) represents the destination state of transition \(t_i\). Note that the states \((s_0,s_1,\cdots,s_m)\) are not necessarily distinct, reflecting potential loops or revisits within the transition path.

\par \textbf{Design of primary function}: Modeling an application begins with defining its primary functions, such as the photo-taking feature in a camera app or the call functionality in a contacts app. By expanding this primary function set, designers can ensure that the EFSM captures a broader range of the application's capabilities. Moreover, a more fine-grained definition of primary functions helps generate more detailed and precise plans, ultimately improving the agent’s performance in executing complex tasks.

\par \textbf{Design of Events}: The event \(e \in E\) is manually designed to instruct the agent on how to perform a specific transition. Designers can use natural language to describe a sequence of operations within a single event. If the event appears in the execution path, its content—after polish—will be incorporated into the final plan. This design provides significant flexibility in the modeling process, allowing complex or tedious interactions to be effectively embedded within the plan.

\textbf{Three Types of Transitions}: Transitions can be broadly categorized into three types based on their functional roles. The first type involves a simple navigation from state \(s\) to \(s'\), where both the action \(a\) and the variable update function \(u(V)\) may be null. The second type corresponds to configuration adjustments within the application, primarily involving updates to the internal variables via \(u(V)\); in this case, the action \(a\) may be null, and the states \(s\) and \(s'\) may remain the same. The third type represents the execution of a primary function, where \(s\) and \(s'\) are often identical, and \(u(V)\) is typically null.

%***********************Algorithm 1***********************

\begin{algorithm*}[t] 
\caption{Workflow of SPlanner}
\label{al:(1)}
\begin{algorithmic}
\State \textbf{Application Modeling:} Use EFSM to model all target applications and obtain the EFSM set \(\mathcal{F}\) as defined in \eqnref{eq:(2)}.
\Statex
\State \textbf{Plan Generation:} Given user instruction \(I\) and EFSM set \(\mathcal{F}\),
\State \quad 1. Use an LLM to parse \(I\), producing \([\varepsilon_1, \varepsilon_2, \cdots, \varepsilon_j]\) and \([A^T_1, A^T_2, \cdots, A^T_j]\), as shown in \eqnref{eq:(5)}.
\State \quad 2. Use a BFS-based solver to compute the execution paths \(p_i = BFS(\varepsilon_i, A^T_i)\) for each app, and aggregate them into the global path plan \(P = (p_1, p_2, \cdots, p_j)\), as shown in \eqnref{eq:(6)}.
\State \quad 3. Use an LLM to combine \(I\) and \(P\) to generate the final natural language plan.

\Statex
\State \textbf{Task Execution:} Given initial GUI screenshot \(S_0\), instruction \(I\), generated plan \(Plan\), initial action history \(H_0 = \emptyset\), and step counter \(i = 1\),
\While{task is not completed \textbf{and} step limit not reached}
    \State Generate the next operation instruction \(O_i = VLM(I, S_i, Plan, H_i)\).
    \State Update history: \(H_{i+1} = H_i + O_i\).
    \State Increment step: \(i \gets i + 1\).
\EndWhile

\end{algorithmic}
\end{algorithm*}

%***********************Section 3-2***********************

\subsection{EFSM-based Planning Module}
\par Given a user instruction, the SPlanner generates an execution plan in natural language. The generation process consists of three stages, namely instruction parsing, EFSM solving and path polishing.
\par In the first stage—\textbf{instruction parsing}, we use a LLM to extract the target applications as well as the sequence of target primary functions from the user instruction. This process can be formally expressed as
\begin{equation}
    LLM(I)\to ((\varepsilon_1,\varepsilon _2,\cdots,\varepsilon _j),(A^T_1,A_2^T,\cdots,A_j^T)). \label{eq:(5)}
\end{equation}
where  \(I\) denotes the user instruction, \(\varepsilon_j\) is one of the EFSMs corresponding to the applications required to complete the task as defined in \eqnref{eq:(2)}, and \(A^T_j \subseteq A^j\) represents the sequence of target primary functions parsed from the instruction as defined in \eqnref{eq:(3)}.
\par Then, in the second stage—\textbf{EFSM solving}, SPlanner employs a BFS-based state machine solver to derive an execution path that sequentially traverses all target primary functions in each \(A^T\). This process is formally defined as:
\begin{equation}
\label{eq:(6)}
\begin{split}
    &BFS(\varepsilon _i,A^T_i)\to p_i,\\
    &P=(p_1,p_2,\cdots,p_j),\\
    &i=1,2,\cdots,j.
\end{split}
\end{equation}
where \(P\) denotes the entire execution path, and \(p_i\) is the segment derived from the \(i\)-th application, as defined in \eqnref{eq:(4)}. If no valid execution path is found (i.e., \(P = \emptyset\)), the planning module skips the path polishing step and directly returns a fallback plan with the context message: “No feasible execution path exists.”.
\par In the final step—\textbf{path polishing}, SPlanner leverages a LLM, guided by the user instruction, to refine the raw path \(P\) into a coherent execution plan in natural language, composed of a series of steps. 
Compared to the original execution path sequence, the polished plan is concise, actionable, and aligned with human understanding. Additionally, the LLM can enrich the plan with contextual information inferred from the instruction. The final plan is then passed to a VLM for step-by-step execution.

%***********************Section 3-3***********************

\subsection{Plan Execution}
\par SPlanner employs a VLM as the executor to carry out the task step by step. At each step, the VLM takes the user instruction, the current screenshot, the plan generated by the plan module and the history of previously actions as input, and then generates the next action based on the action space. The action space is determined by the benchmark or the operating environment, and typically includes operations such as clicking or long pressing a specified pixel, entering specified text, swiping the screen, etc. After an action is executed within the operating environment, the VLM proceeds to generate subsequent actions until the task is successfully completed or the maximum number of steps is reached.
\par During this process, each step in the plan corresponds to one or more executed actions. To enhance the guiding effect of the plan on the model, we incorporate Chain-of-Thought (CoT) \cite{CoT} prompting, encouraging the model to reason about the current plan step before generating each action. The workflow of SPlanner is summarized in \alref{al:(1)}.

%***********************Figure: Success Rate***********************
\begin{figure*}
  \centering
  \includegraphics[width=1\linewidth]{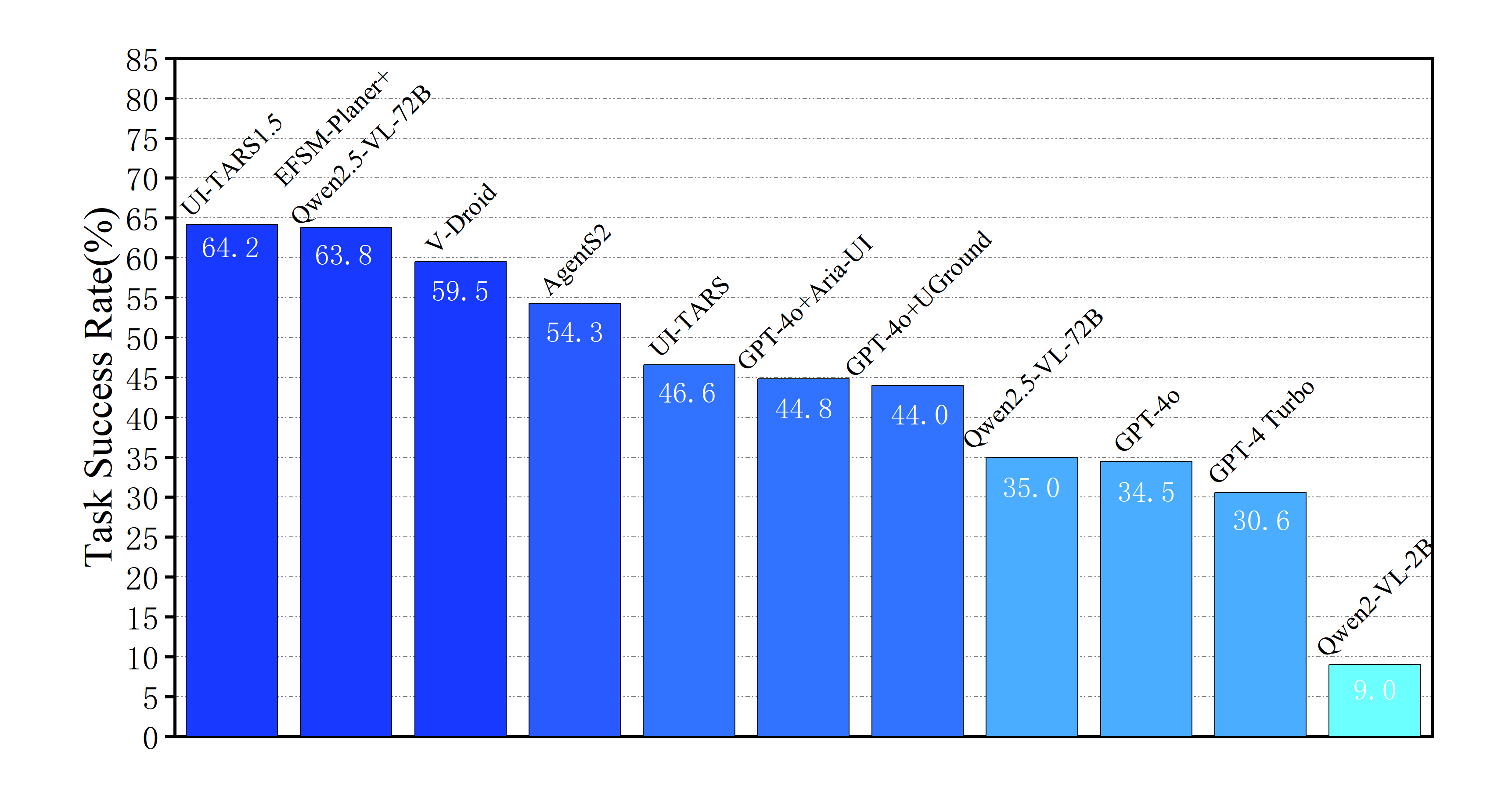}
  \caption{Task success rates of SPlanner and baseline methods on AndroidWorld. For clarity of presentation, darker colors are used to indicate higher success rates, and the exact values are annotated on the corresponding bars.}
  \label{fig:SR_data}
\end{figure*}

%***********************4.0 Experiment***********************
\section{Experiment}
\par In this section, we evaluate the proposed SPlanner on the dynamic mobile GUI agent benchmark AndroidWorld \cite{androidworld} and compare its performance with state-of-the-art (SOTA) methods.

%***********************4.1 Experiment: Benchmark***********************

\subsection{Benchmark}
\par We evaluated our approach on a dynamic mobile GUI agent benchmark that simulates various real-world tasks within a sandbox environment (e.g., a mobile phone simulator). In this setting, the agent receives natural language instructions and interacts with the simulated GUI by executing operations from a predefined set. The benchmark provides user instructions and requires the agent to achieve a specific goal within a limited number of steps. During execution, the agent is free to choose its path, as the benchmark does not impose restrictions on intermediate decisions. Compared with static datasets(e.g. AndroidControl \cite{androidcontrol}), dynamic benchmarks better reflect real-world scenarios and pose greater challenges for agent reasoning and planning.
\par \textbf{AndroidWorld} is a dynamic benchmark designed for evaluating mobile GUI agents. It includes 116 tasks across 20 real-world mobile applications, with task difficulty ranging from fewer than ten steps to over thirty steps. The benchmark encompasses a broad spectrum of scenarios, such as app-specific question answering, cross-application interactions, and in-app content editing. Some of the more challenging tasks require complex visual reasoning—such as extracting relevant information from lengthy on-screen text. Additionally, AndroidWorld imposes strict step limits for each task, significantly reducing fault tolerance and emphasizing the importance of precise and efficient planning..

%***********************4.2 Experiment: Baselines and Implementation Details***********************

\subsection{Baselines and Implementation Details}

\textbf{Comparison Baselines}: We compare SPlanner with several representative baselines on the AndroidWorld benchmark. These include general GUI agents such as UI-TARS \cite{UI-TARS} and AgentS2 \cite{AgentS2}, mobile-specific GUI agents like V-Droid \cite{V-Droid}, and state-of-the-art vision-language models including GPT-4o \cite{gpt4o} and Qwen2.5-VL-72B \cite{qwen2.5vl}. We also include composite methods such as GPT-4o+Aria-UI \cite{aria-ui} and GPT-4o+UGround \cite{Uground}. All comparison results are obtained from the original papers or the official benchmark repositories associated with each method.

\textbf{Implementation Details}: For the SPlanner, we use Deepseek V3 \cite{deepseekv3} to handle instruction parsing and path polishing in the planning module and the execution plan generated by SPlanner is directly included in the VLM’s prompt, along with simple prompt design techniques to encourage step-by-step reasoning in the style of Chain-of-Thought. As the executor, we adopt Qwen2.5-VL-72B as the VLM. Notably, SPlanner does not rely on any fine-tuning or self-evolution techniques—the large models are used in their open-source, general-purpose form without task-specific adaptation.

%***********************4.3 Experiment: Result Comparison***********************
\subsection{Result Comparison}
As shown in \figref{fig:SR_data}, our method (denoted as EFSM-Planner + Qwen2.5-VL-72B) achieved a task success rate of 63.8\%, representing a substantial improvement of 28.8 percentage points over the baseline Qwen2.5-VL-72B (35.0\%). Compared with other advanced approaches, our method outperformed AgentS2 (54.3\%) by 9.5 percentage points and V-Droid (59.5\%) by 4.3 percentage points. Although slightly lower than the current state-of-the-art method UI-TARS1.5 (64.2\%), the performance of our approach demonstrates strong competitiveness, particularly given its plug-and-play nature.

%***********************5.0 Discussion***********************
\section{Discussion}
\par \textbf{Effectiveness of the Proposed Planning Module}: The experimental results demonstrate that SPlanner significantly improves the task completion capabilities of GUI agents, yielding a 28.8 percentage point increase over the baseline VLM (Qwen2.5-VL-72B) on the AndroidWorld benchmark. Our observations suggest that the plans generated by SPlanner effectively reduce instances where the VLM gets "lost" within the screens, and they also enable the execution of fine-grained and complex operations. These capabilities contribute substantially to the improved task success rate. The notable performance gain underscores the effectiveness of integrating structured knowledge into the planning process. By modeling application behavior using EFSMs, SPlanner generates interpretable and reliable execution plans that compensate for the limited multi-step reasoning abilities of current VLMs.

\par \textbf{Plug-and-Play Flexibility}: SPlanner requires only the user’s instruction to generate an execution plan, which is expressed in natural language. This plan can be seamlessly integrated into the input of the executor as part of the prompt, without the need for model fine-tuning or architectural modifications. As a result, SPlanner is not limited to vision-language models (VLMs) but can also be applied to text-only LLM-based GUI agents. This plug-and-play design greatly enhances the versatility and ease of deployment of SPlanner across different types of agents and systems, making it a practical solution for real-world mobile GUI automation.

\par \textbf{Analyzing Failures in Task Execution}: During the experiments, we observed that although EFSM-Planner was able to generate correct plans for the vast majority of tasks, the agent still failed to complete a significant portion of them successfully. Based on our analysis, there are three primary factors contributing to this discrepancy. First, the visual language model (VLM) does not always adhere strictly to the given plan. It may execute actions not specified in the plan or skip planned steps based on its own internal preferences, likely acquired during fine-tuning. Second, certain tasks impose high demands on the VLM’s visual understanding capabilities—for instance, interpreting complex images or comprehending large volumes of text. These challenges cannot be addressed solely by providing a high-level plan. Third, even when the plan is logically correct, it may lack the necessary precision for complex tasks. For example, in a task requiring the deletion of redundant expenses while retaining one entry per category, it is impossible to predefine exact steps. In such cases, the EFSM-Planner can only guide the agent to the appropriate interface and provide general instructions such as “Long-press to select redundant entries of the same type and tap the trash icon in the upper-right corner to delete them.” However, plans of this nature may not be sufficiently specific for the agent to complete the task reliably.

\par \textbf{Consumption of Application Modeling via EFSM}: Currently, building EFSMs for applications involves manual effort. Modelers must be well-acquainted with the application's structure and operational logic to carefully define key EFSM components—particularly events and primary functions—so that SPlanner can reliably interpret user instructions and generate accurate plans. The modeling process typically takes one to two hours per application, with more time required for complex apps featuring intricate workflows or extensive functionality to ensure sufficient coverage. Automating EFSM construction remains a significant challenge. In future work, we aim to explore AI-assisted techniques to streamline this process and reduce the dependence on manual labor.

\section{Conclusion}
\par In this paper, we propose SPlanner, an EFSM-based planning module designed to stably generate task execution plans for GUI agents. SPlanner leverages EFSM to model mobile applications, building a structured knowledge base that supports effective planning. During task execution, SPlanner first parses user instructions, then solves the corresponding EFSM to derive an execution path, which is subsequently refined into a clear and actionable plan. This plan is incorporated into the prompt to guide the VLM executor in generating interaction commands. We evaluate SPlanner on the dynamic benchmark AndroidWorld, and experimental results demonstrate that it significantly improves the task success rate of existing generalist models, verifying its effectiveness. However, there remain challenges in the collaboration between SPlanner and current VLMs—particularly the VLM’s incomplete adherence to the provided plans—which partially limits the overall performance. Improving this synergy will be a key focus of future work. Additionally, since constructing EFSMs currently relies heavily on manual modeling and expert knowledge, we plan to explore automatic EFSM generation methods to enhance the scalability and practicality of SPlanner.

\section{Limitations}
One limitation of SPlanner lies in its reliance on manual modeling for each target application, which incurs significant development cost and requires prior experience with EFSM design. Moreover, the primary functions must be described in precise and unambiguous natural language during the modeling process; otherwise, SPlanner may fail to correctly parse user instructions, leading to inaccurate execution plans. This places a high burden on modelers and limits the system’s scalability in practical deployment. Future work will focus on automating the EFSM construction process and improving the robustness of instruction parsing to reduce modeling effort and enhance scalability.

\bibliography{custom}

\begin{thebibliography}{22}
\providecommand{\natexlab}[1]{#1}

\bibitem[{Agashe et~al.(2025)Agashe, Wong, Tu, Yang, Li, and Wang}]{AgentS2}
Saaket Agashe, Kyle Wong, Vincent Tu, Jiachen Yang, Ang Li, and Xin~Eric Wang.
  2025.
\newblock Agent s2: A compositional generalist-specialist framework for
  computer use agents.
\newblock \emph{arXiv preprint arXiv:2504.00906}.

\bibitem[{Bai et~al.(2025)Bai, Chen, Liu, Wang, Ge, Song, Dang, Wang, Wang,
  Tang et~al.}]{qwen2.5vl}
Shuai Bai, Keqin Chen, Xuejing Liu, Jialin Wang, Wenbin Ge, Sibo Song, Kai
  Dang, Peng Wang, Shijie Wang, Jun Tang, and 1 others. 2025.
\newblock Qwen2. 5-vl technical report.
\newblock \emph{arXiv preprint arXiv:2502.13923}.

\bibitem[{Cheng et~al.(2024)Cheng, Sun, Chu, Xu, Li, Zhang, and Wu}]{seeclick}
Kanzhi Cheng, Qiushi Sun, Yougang Chu, Fangzhi Xu, Yantao Li, Jianbing Zhang,
  and Zhiyong Wu. 2024.
\newblock \href {https://arxiv.org/abs/2401.10935} {Seeclick: Harnessing gui
  grounding for advanced visual gui agents}.
\newblock \emph{Preprint}, arXiv:2401.10935.

\bibitem[{Dai et~al.(2025)Dai, Jiang, Cao, Li, Yang, Tan, Li, and
  Qiu}]{V-Droid}
Gaole Dai, Shiqi Jiang, Ting Cao, Yuanchun Li, Yuqing Yang, Rui Tan, Mo~Li, and
  Lili Qiu. 2025.
\newblock Advancing mobile gui agents: A verifier-driven approach to practical
  deployment.
\newblock \emph{arXiv preprint arXiv:2503.15937}.

\bibitem[{Gou et~al.(2024)Gou, Wang, Zheng, Xie, Chang, Shu, Sun, and
  Su}]{Uground}
Boyu Gou, Ruohan Wang, Boyuan Zheng, Yanan Xie, Cheng Chang, Yiheng Shu, Huan
  Sun, and Yu~Su. 2024.
\newblock Navigating the digital world as humans do: Universal visual grounding
  for gui agents.
\newblock \emph{arXiv preprint arXiv:2410.05243}.

\bibitem[{Huang et~al.(2024)Huang, Liu, Chen, Wang, Wang, Lian, Wang, Tang, and
  Chen}]{planning_servey}
Xu~Huang, Weiwen Liu, Xiaolong Chen, Xingmei Wang, Hao Wang, Defu Lian, Yasheng
  Wang, Ruiming Tang, and Enhong Chen. 2024.
\newblock \href {https://arxiv.org/abs/2402.02716} {Understanding the planning
  of llm agents: A survey}.
\newblock \emph{Preprint}, arXiv:2402.02716.

\bibitem[{Hurst et~al.(2024)Hurst, Lerer, Goucher, Perelman, Ramesh, Clark,
  Ostrow, Welihinda, Hayes, Radford et~al.}]{gpt4o}
Aaron Hurst, Adam Lerer, Adam~P Goucher, Adam Perelman, Aditya Ramesh, Aidan
  Clark, AJ~Ostrow, Akila Welihinda, Alan Hayes, Alec Radford, and 1 others.
  2024.
\newblock Gpt-4o system card.
\newblock \emph{arXiv preprint arXiv:2410.21276}.

\bibitem[{Lee et~al.(2024)Lee, Choi, Lee, Wasi, Choi, Ko, Oh, and
  Shin}]{mobilegpt}
Sunjae Lee, Junyoung Choi, Jungjae Lee, Munim~Hasan Wasi, Hojun Choi, Steve Ko,
  Sangeun Oh, and Insik Shin. 2024.
\newblock Mobilegpt: Augmenting llm with human-like app memory for mobile task
  automation.
\newblock In \emph{Proceedings of the 30th Annual International Conference on
  Mobile Computing and Networking}, pages 1119--1133.

\bibitem[{Li et~al.(2024)Li, Bishop, Li, Rawles, Campbell-Ajala, Tyamagundlu,
  and Riva}]{androidcontrol}
Wei Li, William Bishop, Alice Li, Chris Rawles, Folawiyo Campbell-Ajala, Divya
  Tyamagundlu, and Oriana Riva. 2024.
\newblock On the effects of data scale on computer control agents.
\newblock \emph{arXiv e-prints}, pages arXiv--2406.

\bibitem[{Liu et~al.(2024)Liu, Feng, Xue, Wang, Wu, Lu, Zhao, Deng, Zhang, Ruan
  et~al.}]{deepseekv3}
Aixin Liu, Bei Feng, Bing Xue, Bingxuan Wang, Bochao Wu, Chengda Lu, Chenggang
  Zhao, Chengqi Deng, Chenyu Zhang, Chong Ruan, and 1 others. 2024.
\newblock Deepseek-v3 technical report.
\newblock \emph{arXiv preprint arXiv:2412.19437}.

\bibitem[{Liu et~al.(2023)Liu, Jiang, Zhang, Liu, Zhang, Biswas, and
  Stone}]{llm+p}
Bo~Liu, Yuqian Jiang, Xiaohan Zhang, Qiang Liu, Shiqi Zhang, Joydeep Biswas,
  and Peter Stone. 2023.
\newblock Llm+ p: Empowering large language models with optimal planning
  proficiency.
\newblock \emph{arXiv preprint arXiv:2304.11477}.

\bibitem[{Ma et~al.(2024)Ma, Zhang, and Zhao}]{coco-agent}
Xinbei Ma, Zhuosheng Zhang, and Hai Zhao. 2024.
\newblock Coco-agent: A comprehensive cognitive mllm agent for smartphone gui
  automation.
\newblock \emph{arXiv preprint arXiv:2402.11941}.

\bibitem[{Qin et~al.(2025)Qin, Ye, Fang, Wang, Liang, Tian, Zhang, Li, Li,
  Huang et~al.}]{UI-TARS}
Yujia Qin, Yining Ye, Junjie Fang, Haoming Wang, Shihao Liang, Shizuo Tian,
  Junda Zhang, Jiahao Li, Yunxin Li, Shijue Huang, and 1 others. 2025.
\newblock Ui-tars: Pioneering automated gui interaction with native agents.
\newblock \emph{arXiv preprint arXiv:2501.12326}.

\bibitem[{Rawles et~al.(2024)Rawles, Clinckemaillie, Chang, Waltz, Lau, Fair,
  Li, Bishop, Li, Campbell-Ajala et~al.}]{androidworld}
Christopher Rawles, Sarah Clinckemaillie, Yifan Chang, Jonathan Waltz,
  Gabrielle Lau, Marybeth Fair, Alice Li, William Bishop, Wei Li, Folawiyo
  Campbell-Ajala, and 1 others. 2024.
\newblock Androidworld: A dynamic benchmarking environment for autonomous
  agents.
\newblock \emph{arXiv preprint arXiv:2405.14573}.

\bibitem[{Wei et~al.(2022)Wei, Wang, Schuurmans, Bosma, Xia, Chi, Le, Zhou
  et~al.}]{CoT}
Jason Wei, Xuezhi Wang, Dale Schuurmans, Maarten Bosma, Fei Xia, Ed~Chi, Quoc~V
  Le, Denny Zhou, and 1 others. 2022.
\newblock Chain-of-thought prompting elicits reasoning in large language
  models.
\newblock \emph{Advances in neural information processing systems},
  35:24824--24837.

\bibitem[{Wen et~al.(2024)Wen, Li, Liu, Zhao, Yu, Li, Jiang, Liu, Zhang, and
  Liu}]{autodroid}
Hao Wen, Yuanchun Li, Guohong Liu, Shanhui Zhao, Tao Yu, Toby Jia-Jun Li, Shiqi
  Jiang, Yunhao Liu, Yaqin Zhang, and Yunxin Liu. 2024.
\newblock \href {https://arxiv.org/abs/2308.15272} {Autodroid: Llm-powered task
  automation in android}.
\newblock \emph{Preprint}, arXiv:2308.15272.

\bibitem[{Xu et~al.(2024)Xu, Liu, Sun, Cheng, Yu, Lai, Zhang, Zhang, Tang, and
  Dong}]{androidlab}
Yifan Xu, Xiao Liu, Xueqiao Sun, Siyi Cheng, Hao Yu, Hanyu Lai, Shudan Zhang,
  Dan Zhang, Jie Tang, and Yuxiao Dong. 2024.
\newblock Androidlab: Training and systematic benchmarking of android
  autonomous agents.
\newblock \emph{arXiv preprint arXiv:2410.24024}.

\bibitem[{Yang et~al.(2024)Yang, Wang, Li, Luo, Chen, Huang, and Li}]{aria-ui}
Yuhao Yang, Yue Wang, Dongxu Li, Ziyang Luo, Bei Chen, Chao Huang, and Junnan
  Li. 2024.
\newblock Aria-ui: Visual grounding for gui instructions.
\newblock \emph{arXiv preprint arXiv:2412.16256}.

\bibitem[{Yang et~al.(2023)Yang, Ishay, and Lee}]{llm+asp}
Zhun Yang, Adam Ishay, and Joohyung Lee. 2023.
\newblock Coupling large language models with logic programming for robust and
  general reasoning from text.
\newblock \emph{arXiv preprint arXiv:2307.07696}.

\bibitem[{Zhang et~al.(2024{\natexlab{a}})Zhang, He, Qian, Li, Li, Qin, Kang,
  Ma, Liu, Lin et~al.}]{GUI_agent_survey}
Chaoyun Zhang, Shilin He, Jiaxu Qian, Bowen Li, Liqun Li, Si~Qin, Yu~Kang,
  Minghua Ma, Guyue Liu, Qingwei Lin, and 1 others. 2024{\natexlab{a}}.
\newblock Large language model-brained gui agents: A survey.
\newblock \emph{arXiv preprint arXiv:2411.18279}.

\bibitem[{Zhang et~al.(2024{\natexlab{b}})Zhang, Wu, Teng, Liao, Xu, Xiao, Wei,
  and Tang}]{coat}
Jiwen Zhang, Jihao Wu, Yihua Teng, Minghui Liao, Nuo Xu, Xiao Xiao, Zhongyu
  Wei, and Duyu Tang. 2024{\natexlab{b}}.
\newblock Android in the zoo: Chain-of-action-thought for gui agents.
\newblock \emph{arXiv preprint arXiv:2403.02713}.

\bibitem[{Zheng et~al.(2024)Zheng, Gou, Kil, Sun, and Su}]{seeact}
Boyuan Zheng, Boyu Gou, Jihyung Kil, Huan Sun, and Yu~Su. 2024.
\newblock Gpt-4v (ision) is a generalist web agent, if grounded.
\newblock \emph{arXiv preprint arXiv:2401.01614}.

\end{thebibliography}
\nocite{*}
\label{sec:appendix}

\end{document}